\documentclass{article}

\usepackage[preprint]{iaseai26}

\makeatother

\usepackage[utf8]{inputenc} 
\usepackage[T1]{fontenc}    
\usepackage{hyperref}       
\usepackage{url}            
\usepackage{booktabs}       
\usepackage{amsfonts}       
\usepackage{nicefrac}       
\usepackage{microtype}      
\usepackage{xcolor}         
\usepackage{natbib}
\usepackage{tabularx}
\usepackage{graphicx}
\usepackage{tikz}
\usetikzlibrary{shapes,arrows.meta,positioning}
 \usepackage{amsmath}

\title{Interpreting Agentic Systems: Beyond Model Explanations to System-Level Accountability}

\author{
  Judy Zhu* \\
  Vector Institute for Artificial Intelligence \\
  \texttt{judy.zhu@vectorinstitute.ai} \\
  \And
  Dhari Gandh* \\
  Vector Institute for Artificial Intelligence \\
  \texttt{dhari.gandhi@vectorinstitute.ai} \\
  \And
  Himanshu Joshi \\
  University of Texas, Austin  \\
  \texttt{himanshujoshi@utexas.edu} \\
  \And
  Ahmad Rezaie Mianroodi \\
  Dalhousie University \& Vector Institute for Artificial Intelligence \\
  \texttt{ahmad.rm@dal.ca} \\
  \And
  Sedef Akinli Kocak \\
  Vector Institute for Artificial Intelligence \\
  \texttt{sedef.kocak@vectorinstitute.ai} \\
  \And
  Dhanesh Ramachandran \\
  Vector Institute for Artificial Intelligence \\
  \texttt{dhanesh.ramachandran@vectorinstitute.ai} \\
}

\begin{document}
\thanks{* These authors contributed equally to this work.}

\maketitle

\begin{abstract}
    Agentic systems have transformed how Large Language Models (LLMs) can be leveraged to create autonomous systems with goal-directed behaviors, consisting of  multi-step planning and the ability to interact with different environments. These systems differ fundamentally from traditional machine learning models, both in architecture and deployment, introducing unique AI safety challenges, including goal misalignment, compounding decision errors, and coordination risks among interacting agents, that necessitate embedding interpretability and explainability by design to ensure traceability and accountability across their autonomous behaviors. Current interpretability techniques, developed primarily for static models, show limitations when applied to agentic systems. The temporal dynamics, compounding decisions, and context-dependent behaviors of agentic systems demand new analytical approaches. This paper assesses the suitability and limitations of existing interpretability methods in the context of agentic systems, identifying gaps in their capacity to provide meaningful insight into agent decision-making. We propose future directions for developing interpretability techniques specifically designed for agentic systems, pinpointing where interpretability is required to embed oversight mechanisms across the agent lifecycle from goal formation, through environmental interaction, to outcome evaluation. These advances are essential to ensure the safe and accountable deployment of agentic AI systems. 

\end{abstract}

\section{Introduction}

The emergence of generative artificial intelligence (AI), particularly Large Language Models (LLMs), has created unprecedented opportunities and challenges for modern society. ChatGPT achieved the fastest adoption rate of any consumer technology in computing history upon its launch~\citep{hu2023chatgpt}. The proliferation of applications and research in generative AI and LLMs has accelerated dramatically in recent years. As society continues adapting to generative AI's impact, AI researchers are already advancing toward the next evolutionary stage: agentic systems. The term ‘agentic AI’, which was minimally searched on Google Trends prior to April 2024, has since been climbing to its peak in July 2025~\citep{futureinternet}. This meteoric rise has created a discrepancy between new innovations' promising potential and the capabilities of trustworthy real world applications of agentic systems.

Agentic systems represent an extension of LLM capabilities, where one or multiple agents autonomously plan, collaborate, review, and make decisions to achieve defined objectives. These systems move beyond single-query responses to orchestrate complex, multi-step workflows. The evolution of AI coding assistants illustrates this transition. Early LLM-based coding assistants like ChatGPT~\citep{chatgpt},  and Claude~\citep{Claude} operated primarily as consultative tools where users would paste code snippets or describe problems, and the model would provide suggestions or explanations within a single interaction. These systems lacked persistent memory across sessions and could autonomously navigate or modify codebases. Similarly, initial versions of GitHub Copilot~\citep{copilot} functioned mainly as intelligent autocomplete, suggesting code completions based on local context without broader project understanding.
Resent agentic coding assistants such as Claude Code~\citep{claudecode} demonstrate qualitatively different capabilities. These systems can autonomously traverse entire codebases, search external documentation, formulate multi-step implementation plans, edit multiple files across a project, and execute code within the user's local environment, all while maintaining context and adapting based on execution results.

The development of generative AI-powered coding assistants represents a prominent application demonstrating the capabilities of agentic systems. However, the autonomous nature of these systems has prompted significant concerns within the AI research community regarding their safety in terms of transparency and trustworthiness~\citep{murugesan2025rise}, revealing a need for greater interpretability. As agentic systems gain increasingly sophisticated decision-making capabilities and environmental interaction privileges, establishing robust mechanisms for explainability, oversight, and accountability has become a critical research priority. 

In this paper, we examine interpretability as foundational to trust and safety in agentic systems, and outline future research directions aimed at advancing methods for transparent, accountable, and reliable system behavior. 
The remainder of this paper is organized as follows: Section~\ref{Sec:Background} provides foundational background on agentic systems, including definitions, distinctions, taxonomies, and core architectures. Section~\ref{Sec:LitReview} reviews the evolution towards agentic systems, summarizes state-of-the-art techniques, highlights real-world examples, and identifies major challanges related to opacity, planning capabilities, and autonomy. Section~\ref{Sec:Interpretability} examines  the motivation and requirements for interpretability in agentic systems, assessing the limitations of existing techniques and their applicability to these architectures. Section~\ref{Sec:Future} proposes future research directions for developing interpretability techniques specifically designed for safe and accountable agentic systems. Finally, Section \ref{Sec:Conclusion}, concludes the paper and outlines key insights.

\section{Background: AI Agents vs Agentic Systems}
\label{Sec:Background}
A clear understanding of agentic systems is essential to contextualize the interpretability challenges explored in this paper. As agentic systems increasingly form autonomous decision-making capabilities across domains, it becomes increasingly important to understand their fundamental and functional foundations. Establishing this foundation clarifies how agentic systems differ from traditional AI agents and highlights where interpretability gaps begin to emerge.

\subsection{Definitions}
An AI Agent is a sophisticated system designed for autonomous goal achievement, using a LLM as its core reasoning engine rather than being an LLM itself~\citep{sapkota2025ai, derouiche2025agentic}. The LLM acts as the ``brain'', providing capabilities for understanding, planning, and generating natural language~\citep{glanois2024survey}. The agent, however, orchestrates a broader set of functionalities, including perception, memory, tool utilization, and an iterative action-feedback loop to interact with and effect changes in its environment~\citep{derouiche2025agentic, sapkota2025ai}. It uses the LLM for its cognitive capacity and provides the framework and mechanisms to operationalize this comprehension and intelligence in a dynamic environment~\citep{futureinternet}. This allows AI agents to strategize and tackle complex, multi-step tasks that go beyond simple text generation or question answering~\citep{wei2022chain, wang2024survey}.

Agentic systems represent a drastic paradigm shift from traditional AI agents and their applications. While AI agents are modular architectures powered by LLMs for task-specific automation, agentic systems are complex, multi-agent systems characterized by emergent behaviors and coordinated autonomy, enabling them to dynamically orchestrate fully fledged workflows~\citep{sapkota2025ai}. These systems consist of specialized sub-agents that collaboratively plan, reason, and coordinate to address problems at a system-level, pursuing shared objectives as opposed to isolated tasks~\citep{futureinternet}. 

Additional fundamental differences between agents and agentic systems are summarized in the \textbf{\textit{Appendix Table \ref{tab:core_identity}}}.

\subsection{Functional Components}

\begin{figure}
    \centering
    \includegraphics[width=0.75\linewidth]{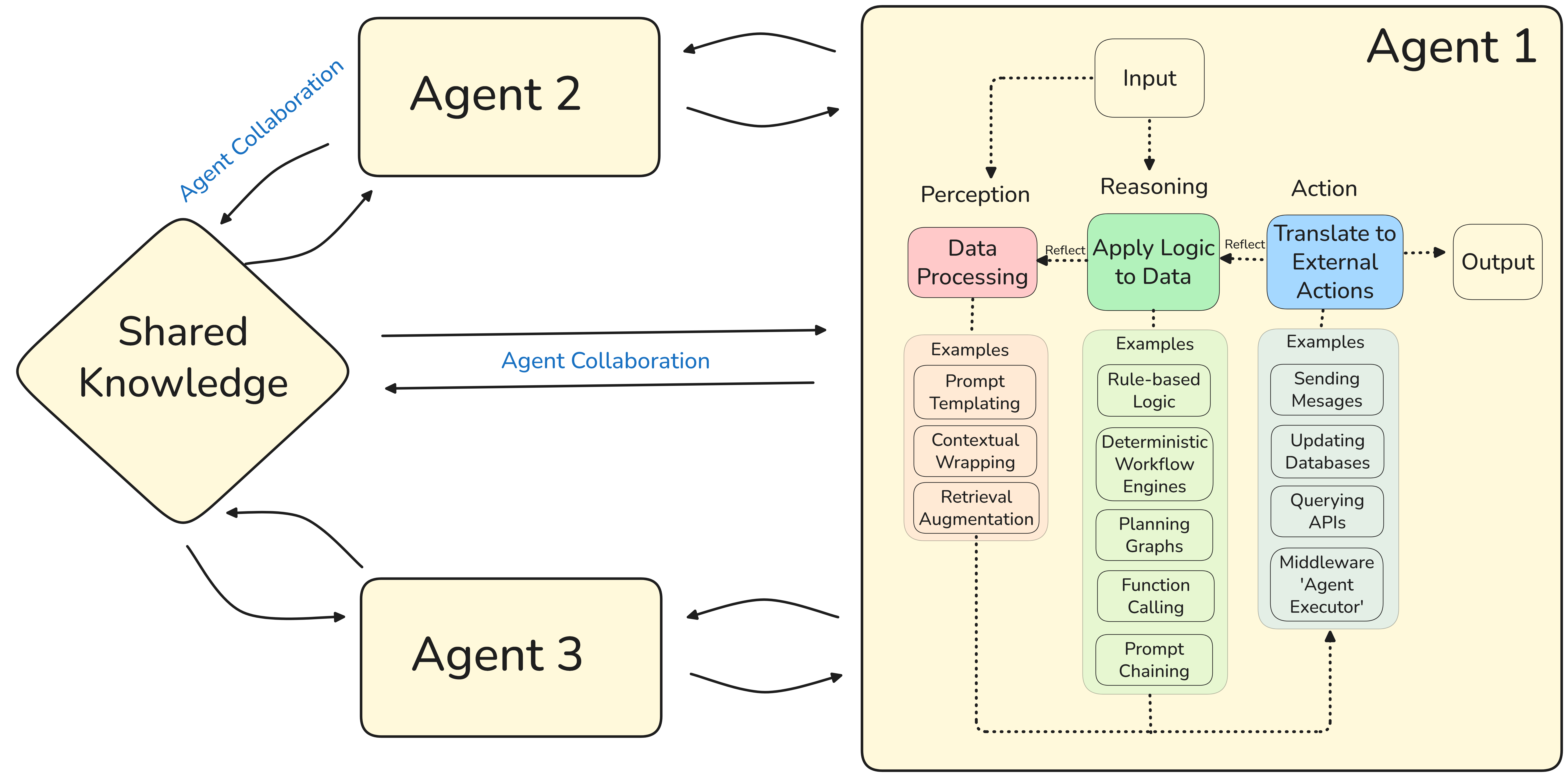}
    \caption{An Agentic System Core Anatomy and Task Examples}
    \label{fig: anatomy}
\end{figure}

There is a distinct shift in modular architectural design from AI agents to agentic systems. As illustrated in Figure~\ref{fig: anatomy}, traditional AI agents are built around three foundational modules: perception, reasoning and action, which enable limited autonomy within well-defined tasks. The perception module processes input data (such as user prompts) and prepares it for the reasoning module. The reasoning module, being the most crucial module, applies logic to the prompt data (through the use of a LLM to interpret context and infer outcomes). Lastly, the action module conveys these inferred decisions by interacting with external systems, tools, or APIs to produce outputs.

Agentic systems build upon this foundation with advanced modules, as seen in Figure~\ref{fig: anatomy}, including: specialized agents (multi-agent collaboration), advanced reasoning and planning (task decomposition), persistent memory (shared contexts), and orchestration (system-wide coordination) \citep{sapkota2025ai, liu2025agentpatterncatalogue}. These additional layers enable distributed intelligence across multiple agents. Each agent performs a defined role while contributing to a shared knowledge base that supports coordination and adaptive learning.


Together, these modules represent the architectural backbone of agentic systems, unifying perception, reasoning, action, and memory under a coordinated autonomy paradigm. After outlining the structural and functional anatomy of agentic systems, it becomes essential to contextualize these architectures within the wider landscape of AI’s technological evolution. Understanding how today's agentic systems evolved from early machine learning paradigms provides important context for their capabilities and limitations, highlighting how interpretability issues have developed into system-level challenges.

\section{Literature Review}
\label{Sec:LitReview}
    \subsection{Historical perspective: Early Machine Learning to Agentic Architectures}
    
    \begin{figure}
        \centering
        \includegraphics[width=1.0\linewidth]{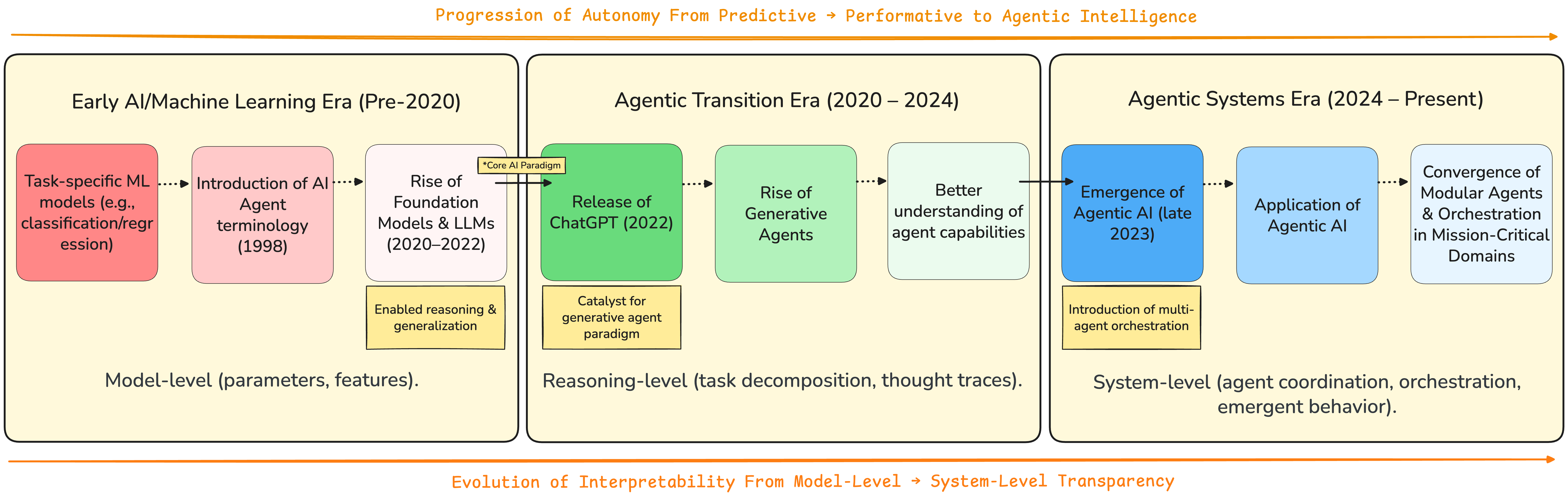}
        \caption{Evolution of AI Agents to Agentic Systems across three eras, illustrating the parallel progression of autonomy and interpretability.}
        \label{fig:evolution}
    \end{figure}

    As illustrated in Figure~\ref{fig:evolution}, the development of artificial intelligence can be broadly divided into three eras: the \textit{Early AI/Machine Learning Era}, the \textit{Agentic Transition Era}, and the \textit{Agentic Systems Era}. Each era represents a dual trajectory: an increase in autonomous capability and a corresponding shift in interpretability focus.
    
    Historically, AI progressed from task-specific machine learning, optimized for classification or regression in static settings, to foundation and generative models that learn broad representations and exhibit emergent compositional reasoning \citep{wei2022chain, glanois2024survey}. LLMs enabled open-ended behavior across varied contexts, establishing the basis for interaction rather than one-off prediction. Building on these capabilities, LLM-based agents introduced mechanisms for planning, memory, and tool use to pursue goals. Overtime these agents started shifting from “predictive” to “performative” AI \citep{wang2024survey}. These agents began demonstrate iterative reasoning and environment-aware decision-making marking early steps toward autonomy. 

    The next phase in this evolution is the emergence of \textit{Agentic Systems}: multi-agent architectures composed of specialized, communicating agents capable of distributed planning, negotiation, and self-organization \cite{sapkota2025ai, derouiche2025agentic}. Unlike agents, agentic systems exhibit coordinated autonomy, reflecting structures akin to human organizations, with meta-agents orchestrating task decomposition and synthesis.

    This progression, from statistical learners to foundation models, then to autonomous agents and multi-agent systems, marks a shift from isolated intelligence to collective cognition. These systems extend AI’s capability frontier, but also simultaneously amplify challenges around transparency, accountability, and interpretability. These are all issues central to the safe deployment of agentic systems.
  
    \subsection{State-of-the-Art Techniques for Agentic Systems}
    The state-of-the-art in agentic systems reflect a convergence of enabling techniques that operationalize the architectural components introduced in the previous section. While large foundation models such as LLMs and multi-modal models provide the cognitive substrate for perception and reasoning, the progression toward \textit{agentic} architectures has required specialized methods that support persistence, coordination, and control. These techniques transform static model capabilities into dynamic system functions, enabling agents not only to reason, but also to remember, plan, and act collaboratively across environments.

    Recent deployments indicate that high-performing agentic systems are designed using the following techniques summarized in Table~\ref{tab:agentic_techniques}. Collectively, these define the technical mechanisms that underpin autonomy and adaptability in real-world settings. 

    \begin{table}[h!]
    \centering
    \small
    \caption{Key techniques enabling core capabilities in Agentic systems.}
    \label{tab:agentic_techniques}
    \renewcommand{\arraystretch}{1.4} 
    \begin{tabularx}{\textwidth}{p{0.2\textwidth} p{0.35\textwidth} X}
    \toprule
    \textbf{Technique Area} & \textbf{Functionality} & \textbf{Implementation Examples} \\
    \midrule
    
    \textbf{Memory and Context Management} & 
    Supports persistence, continuity, and adaptive reasoning through episodic, semantic, and procedural memory layers that maintain both short- and long-term context. &
    Implemented in systems such as \textit{MIRIX} and \textit{MemTool}, which use vector databases to store and retrieve contextual information for long-horizon reasoning~\cite{derouiche2025agentic}. \\
    
    \textbf{Tool Integration and API Invocation} & 
    Extends agentic capabilities beyond pretraining by enabling agents to interact with and control external tools, APIs, and environments. &
    Frameworks like \textit{MATPO} train “planner–executor” agent pairs to select and invoke tools reliably in dynamic, multi-agent settings~\citep{causalsufficiency}. \\
    
    \textbf{Agent Coordination and Role Assignment} & 
    Facilitates collaboration and dynamic role allocation among specialized agents, enhancing scalability and adaptability in complex tasks. &
    Systems such as \textit{AgentFlow} employ orchestrators that monitor context, assign roles, and coordinate inter-agent communication in real time~\citep{derouiche2025agentic, sapkota2025ai}. \\
    
    \textbf{Planning and Task Decomposition} & 
    Enables complex, multi-stage problem-solving through iterative task breakdown, reasoning validation, and re-planning. &
    Implemented through graph-based orchestration frameworks like \textit{LangGraph}, which formalize inter-agent dependencies and task refinement~\citep{sapkota2025ai}. \\
    
    \textbf{Scalability and Efficiency Optimization} & 
    Balances autonomy and computational cost through hierarchical control mechanisms, selective activation, and token-efficient reasoning. &
    Recent architectures integrate multi-layer scheduling and adaptive activation pipelines to optimize reasoning efficiency~\citep{agrawal2025redefining}. \\
    
    \bottomrule
    \end{tabularx}
    \end{table}

    In high-performing deployments, these components operate jointly rather than in isolation. For example, a planner agent decomposes a task using \textit{LangGraph}, stores intermediate outcomes in a shared episodic memory (\textit{MIRIX}), and invokes an external data API via the executor agent (\textit{MATPO}) before final synthesis. Such designs emphasize feedback loops between memory, planning, and execution layers, forming a cohesive reason–act–reflect cycle essential to adaptive autonomy.

     To reflect these design changes, there are new agentic system frameworks, which are either graph-based, workflow oriented, or modular~\citep{liu2025agentpatterncatalogue}. These are reusable libraries and tools that explain how to implement the core agentic architectures. These frameworks include CrewAI, LangGraph, AutoGen, OpenAI Agents SDK, LlamaIndex, MetaGPT, and all need external logic and manual setup for robust reinforcement~\citep{derouiche2025agentic}. More mature and widely adopted frameworks like \textit{AutoGen} and \textit{AutoGPT} have yet to demonstrate deployment readiness in safety-critical or regulated domains such as finance or healthcare, while others like \textit{Google ADK} remain primarily experimental. As highlighted by \citet{derouiche2025agentic} and \citet{sapkota2025ai}, future deployment of agentic frameworks will necessitate the integration of standardized safety and accountability layers, including interpretability modules, guardrail-based orchestration, and verifiable logging to ensure reliability and ethical compliance in critical settings.

    \subsection{Deployment Examples of Agentic Systems}
    
    The emerging adoption of agentic systems across industries demonstrates the early stages translating theoretical advances in autonomy, reasoning, and collaboration into practical, scalable applications. Table~\ref{tab:deployments} summarizes representative examples across domains, highlighting their core components, applications, and outcomes.

    \begin{table}[h!]
    \centering
    \small
    \caption{Representative deployments of agentic systems across domains, highlighting architectures and observed capabilities.}
    \label{tab:deployments}
    \renewcommand{\arraystretch}{1.4} 
    \begin{tabularx}{\textwidth}{p{0.2\textwidth} p{0.22\textwidth} X}
    \toprule
    \textbf{Domain / System} & \textbf{Application Setting} & \textbf{Key Components and Capabilities} \\
    \midrule
    
    \textbf{Minerva CQ} & 
    Customer Service Centers & 
    Real-time agent-assist platform integrating live transcription, intent and sentiment analysis, contextual retrieval, adaptive workflow triggering, and conversational summarization for reduced handling time and improved resolution rates~\citep{agrawal2025redefining}. \\
    
    \textbf{Agentic MRM Crews} & 
    Financial Services & 
    Model risk management “crew” architecture using \textit{CrewAI} and \textit{LangGraph} orchestration. Incorporates Toolformer-style task selection, embedded regulatory constraints, and role-based delegation mirroring human lines of defense~\citep{okpala2025modeling}. \\
    
    \textbf{Agentic UAVs (SAR)} & 
    Autonomous Robotics / Search and Rescue & 
    Five-layer ROS2–Gazebo integrated architecture combining YOLOv11 perception with GPT-4 and Gemma-3 reasoning for dynamic planning and API-driven trajectory adjustment in real-time simulations~\citep{causalsufficiency}. \\
    
    \bottomrule
    \end{tabularx}
    \end{table}

    In addition to industrial deployments, agentic systems have considerable potential to transform the healthcare domain offering opportunities to improve efficiency and reduce time and cost across various operations~\citep{hinostroza2025ai, karunanayake2025next}. Similarly, \citet{elgendy2025agentic} highlight the transformative potential of  agentic systems within the fintech industry. Collectively, these examples illustrate the ongoing maturation of agentic systems from conceptual prototypes to operational deployments across diverse sectors. Each implementation reflects a distinct application of autonomy integrated with context-aware adaptive reasoning. Across these domains a consistent and growing demand is emerging for interpretability, traceability, and controllable autonomy. These examples also reveal the boundaries of present architectures, where decision transparency still remains an open challenge.

    \subsection{Agentic System Risks and Challenges}
    
    As agentic systems expand across sectors, their technical sophistication and operational independence introduce new categories of risk that exceed traditional AI failure modes. These challenges stem from their ability to plan, act, and adapt over long horizons with limited human oversight. The most prominent risks, summarized in Table~\ref{tab:agentic-challenges}, relate to opacity, planning fragility, unchecked autonomy, temporal uncertainty, goal misalignment, and the absence of accountability infrastructure.

    \begin{table}[h!]
    \centering
    \small
    \caption{Fundamental risks and challenges in deploying agentic systems in real-world scenarios.}
    \label{tab:agentic-challenges}
    \renewcommand{\arraystretch}{1.4} 
    \begin{tabularx}{\textwidth}{p{0.21\textwidth} p{0.42\textwidth} X}
    \toprule
    \textbf{Challenge} & \textbf{Description} & \textbf{Implications} \\
    \midrule
    
    \textbf{Opacity of Reasoning} & 
    Internal reasoning processes including goal decomposition, prioritization, and decision paths, remain largely unobservable, making it difficult to trace how agents reach conclusions~\citep{okpala2025modeling, kim2025because}. &
    This lack of transparency undermines auditability and accountability, especially in regulated domains where interpretability is critical for compliance. \\
    
    \textbf{Planning Fragility} & 
    Multi-step and hierarchical reasoning architectures amplify minor biases or utility misestimates, which can cascade through dependent tasks~\citep{okpala2025modeling}. &
    Errors introduced early in the reasoning chain may propagate, causing failures in high-stakes systems such as finance or healthcare. \\
    
    \textbf{Unchecked Autonomy} & 
    Self-initiated agent actions may proceed without human review or constraint, particularly when meta-agents delegate decisions recursively~\cite{okpala2025modeling, kim2025because}. &
    Such autonomy diffuses responsibility and creates a “moral crumple zone,” where accountability is displaced from the system to human overseers after errors occur. \\
    
    \textbf{Temporal Dependencies} & 
    Agentic systems exhibit strong temporal dependencies where earlier decisions influence future reasoning states~\citep{derouiche2025agentic, sapkota2025ai}. &
    The absence of causal traceability complicates counterfactual audits, making it difficult to identify root causes or verify model behavior retrospectively. \\
    
    \textbf{Objective Misalignment} & 
    Reward models or optimization goals often fail to encode tacit human constraints, leading to specification gaming and emergent undesirable behavior~\citep{okpala2025modeling}. &
    Misaligned objectives may result in unethical shortcuts or performance metrics that optimize for efficiency over fairness or safety. \\
    
    \textbf{Lack of Accountability Infrastructure} & 
    Few deployed systems maintain detailed logs of reasoning paths, intermediate states, or inter-agent interactions~\citep{okpala2025modeling, kim2025because}. &
    Without standardized provenance tracking, transparency and governance are severely hindered, limiting the ability to audit or verify system integrity. \\
    
    \bottomrule
    \end{tabularx}
    \end{table}

Taken together, these issues highlight a central vulnerability of the agentic era: as systems increasingly plan, reason, and act over extended horizons, opacity becomes a first-order failure mode rather than a peripheral concern. The cumulative implication is that transparency and interpretability cannot remain optional, after-the-fact diagnostics; they must be treated as foundational design requirements embedded in the architecture and lifecycle of agentic systems.

\section{The Interpretability Imperative for Agentic Systems}
\label{Sec:Interpretability}

    \subsection{Interpretability and Explainability in Traditional ML Models}
    The terms interpretability and explainability are closely related and often used interchangeably. Interpretability generally refers to the ability of a human to understand the cause of a decision or a model’s logic. Explainability usually denotes the techniques or mechanisms through which a model’s workings or predictions can be made understandable \citep{linardatos2020explainable}. 
    
    Broadly, explanation methods fall into two categories: intrinsically interpretable models and post hoc explanation techniques~\citep{linardatos2020explainable,guidotti2018survey,lipton2018mythos,gao2023interpretability,hassija2024interpreting}. Intrinsically interpretable (or self-explaining) approaches design the model itself to be transparent. On the other hand, post hoc methods such as feature attribution techniques or counter-factual explanations attempt to explain a pre-trained ``black-box'' model without altering its internals~\citep{ribeiro2016should,lundberg2017unified,wachter2017counterfactual}.  One can also categorize machine learning interpretability techniques into model-agnostic and model-specific, depending on whether they can be applied to any model type or are tailored to a particular model’s structure. Model-agnostic methods such as LIME \citep{ribeiro2016should} and SHAP \citep{lundberg2017unified} treat models as black boxes and work by analyzing input–output behavior, making them flexible across data modalities like images, text, and tabular data. In contrast, model-specific methods leverage the internal workings of a given architecture to explain its predictions. For example, for convolutional neural networks (CNNs) in vision tasks, techniques like saliency maps \citep{itti2002model} and Grad-CAM \citep{selvaraju2017grad} visualize which image regions most influence predictions. Comprehensive surveys of existing explainability and interpretability techniques for machine learning models are presented in \cite{linardatos2020explainable,gao2023interpretability,hassija2024interpreting}.

 \subsection{Why Current Explainability Methods Fall Short in Agentic Systems?}

Current explainability and interpretability  methods were developed for machine learning models with well-defined mappings between static inputs and outputs, assuming closed-world settings, stable architectures, and localized decision-making. Agentic systems violate these assumptions through dynamic, interdependent modules that reason, plan, and act across time, often modifying their own decision boundaries through interaction.

To illustrate this fundamental incompatibility, consider SHAP values which is a widely used post-hoc interpretability technique. SHAP values assume that features are substitutable and that all $2^n$ coalitions can be meaningfully evaluated to compute marginal contributions. However, in agentic architectures where perception, planning, reasoning, and tool execution form a compositional pipeline with strict dependencies, most coalitions are either undefined or trivially unsuccessful. Evaluating system performance without perception is meaningless when downstream components require perceptual input, violating the assumption that each subset has a well-defined characteristic function value. Moreover, these components exhibit strong complementarityrather than additive contributions and the system performance is multiplicative across components rather than decomposable into independent marginal effects. This architectural non-substitutability renders SHAP values, which excel at explaining contributions among comparable alternatives (e.g., input features to a classifier), unable to meaningfully attribute credit in sequential, tightly-coupled system architectures where components serve irreplaceable functional roles.

\subsection{Motivations for Interpretability in Agentic Systems}

Agentic systems introduce layers of complexity that go beyond traditional, model-centric AI frameworks. Unlike standalone models, they consist of multiple interacting components such as perception, reasoning, memory, planning, and action modules, often distributed across several agents. In a safety-critical scenario, each agent perceives and interprets its environment (e.g., medical readings, sensor logs, natural language descriptions), forms internal beliefs and strategies, exchanges messages with others, and acts based on both internal reasoning and external interactions. The opacity of any of these processes, within or between agents, can compromise the reliability and safety of the system as a whole.


Here, we elaborate on the challenges that may arise within different aspects of agentic systems:

\begin{enumerate}
  \item \textbf{Reasoning and Planning:} With multi-agent distributed reasoning (ReAct, ToT, orchestration layers), reasoning becomes nested and emergent. Who decomposed the goal? Why was a sub-task re-planned? What causal assumptions were made? Traditional interpretability fails because it is local and model-centric, while agentic reasoning is global, system-centric, and unfolds dynamically.

  \item \textbf{Action Selection and Execution:} Execution chains involve multiple agents invoking tools, passing outputs, and retrying failed calls. Debugging why an action was chosen (and who decided it) requires reconstructing nested tool-agent interactions across layers. Current interpretability lacks system-level provenance, making root-cause analysis of unsafe behavior extremely difficult.

  \item \textbf{Memory and State Evolution:} With episodic, semantic, and vector-based persistent memory, there is a lack of visibility into what memories were written, retrieved, or ignored, or how outdated or biased memories shape current reasoning. Interpretability tools don't expose memory dynamics, which is a key gap since oversight often hinges on knowing why an agent acted on past context.

  \item \textbf{Coordination and Communication:} Orchestrators and meta-agents assign roles, manage dependencies, and arbitrate conflicts. Yet there is no interpretability for inter-agent communication: what message was misunderstood, whether agents share context consistently, or whether coordination protocols failed. This absence makes emergent failure modes (deadlocks, loops, divergence from goals) invisible until too late.

  \item \textbf{Emergent Behavior and System-Level Dynamics:} Interpretability for individual models doesn't extend to collective dynamics such as cascading errors across agents, goal misalignment or conflict resolution, and instability when scaling the number of agents. Existing explainability literature lacks tools to explain why the system as a whole behaved a certain way (e.g., why a research automation pipeline drifted into irrelevant exploration).
    
  \item \textbf{Errors and Biases Imposed by Having Humans in the Loop:} As mentioned by \cite{han2023ignorance}, different groups of people find different types of explanations helpful. Laypeople base their trust on explanation faithfulness (how accurately the explanation represents the underlying model) while domain experts rely on explanation alignment (how well explanations match their prior knowledge), suggesting that domain experts may experience cognitive biases due to their expertise. This is a critical aspect of agentic systems that imposes new complexities to explainability. Explanations should be tailored to users while accounting for the biases humans might bring into the system.
\end{enumerate}

It is evident that current post-hoc explainability techniques, originally devised for single-model predictions, are not immediately suitable for agentic systems. They often provide fragmented, local insights (e.g., importance of one feature at one time step) that fail to capture the global, temporal, and interactive aspects of multi-agent systems.

\section{Future Directions for Agentic Interpretability Research}
\label{Sec:Future}

\subsection{The Challenge Ahead}

It is argued here that while existing interpretability techniques remain effective for perception components in agentic systems, fundamental limitations emerge when these systems are examined holistically. Standard techniques such as Testing with Concept Activation Vectors (TCAV) \cite{kim2018interpretability}, integrated gradients, sparse autoencoders, and saliency maps have been shown to work effectively for perception modules built on CNNs, Vision Language Models (VLMs), Small Language Models (SLMs), or LLMs. When perception modules use these architectures, most mechanistic, post-hoc, or intrinsic interpretability methods can be applied without substantial modification~\citep{glanois2024survey, hassija2024interpreting}. The challenge lies not in analyzing individual components, but in understanding the integrated system.

The limitations of component-wise analysis become apparent when system-level failures are investigated. Consider a scenario where an external observer attempts to diagnose why an agentic system failed, armed with saliency maps from perception, attention weights from reasoning, and execution logs from tool use. The critical question becomes: how can this heterogeneous information be aggregated to identify the source of system failure?

This aggregation challenge represents more than an engineering problem. It reflects a fundamental mismatch between the assumptions underlying current interpretability techniques and the failure modes characteristic of agentic systems. Three primary issues can be identified.

First, component-level explanations operate in fundamentally different representational spaces. Saliency maps identify relevant pixels, attention weights highlight salient tokens, and execution logs document performed actions~\citep{selvaraju2017grad,itti2002model,vaswani2017attention,derouiche2025agentic,sapkota2025ai}. These explanation types cannot be directly compared or combined. Determining whether a failure originated from incorrect visual attention versus inappropriate tool selection becomes non-trivial when these are measured using incompatible metrics~\citep{gao2023interpretability,hassija2024interpreting}.

Second, temporal error propagation poses distinct challenges. A perception error at timestep $t$ may not immediately manifest as failure but instead constrain reasoning at $t+2$, influence tool selection at $t+4$, and ultimately produce execution failure at $t+7$~\citep{okpala2025modeling,derouiche2025agentic}. While component-level interpretability can explain why a particular action was selected at a given timestep, it cannot account for why the complete multi-step trajectory failed. Traditional methods assume static, single-pass computations with linear causality. Agentic systems, by contrast, involve iterative feedback loops where outputs become subsequent inputs, tool results modify context, and system state evolves continuously~\citep{sapkota2025ai}.

Third, concurrent execution in production environments introduces additional complexity. When systems process asynchronous requests from multiple users with overlapping execution states, failures generate thousands of component-level explanations. Determining which specific request failed, whether the failure resulted from resource contention or decision errors, and distinguishing correlation from causation across concurrent executions becomes intractable~\citep{derouiche2025agentic,agrawal2025redefining}.

The core limitation is that component-level interpretability implicitly assumes system behavior can be understood through compositional analysis of individual parts. However, agentic systems exhibit emergent failure modes where system-level behavior cannot be reduced to the sum of component-level explanations. What is needed are not improved visualizations of existing explanations, but rather new interpretability frameworks designed for system-level analysis. These frameworks must be capable of tracing causal chains across multi-step executions, tracking error propagation through time, translating between different explanation types, and disambiguating concurrent execution traces.

As traditional explainability techniques struggle to capture the behaviors of agentic systems~\citep{linardatos2020explainable,gao2023interpretability,hassija2024interpreting,derouiche2025agentic,sapkota2025ai}, a complementary line of work reframes interpretability as a human–agent interaction rather than a purely post-hoc artifact. In this view, interpretability is produced through dialogue: the system adapts its explanations to the user’s goals, prior knowledge, and points of confusion over multiple turns. For example, \citet{kim2025because} propose using the LLM as an interactive “teaching” agent that explicitly builds a mental model of the user and uses it to guide multi-turn explanatory exchanges, what they term agentic interpretability: a process that proactively supports human understanding over repeated interactions by modeling the user. At the same time, this shift introduces practical and epistemic difficulties. Because user prompts and follow-up questions become part of the explanatory procedure, the resulting “interpretation” depends on the particular conversational path taken, complicating reproducibility, controlled comparisons, and isolation of causal factors. The joint variability of user behavior and model behavior yields a combinatorial space of trajectories that is hard to benchmark systematically. Moreover, even when an LLM produces fluent rationales, those rationales need not reflect the computations that generated the output. \citet{madsen2401self} show that persuasive reasoning traces can be unfaithful to underlying decision criteria, cautioning against treating generated explanations as reliable evidence of mechanism. Finally, since LLMs can hallucinate~\citep{huang2025survey}, self-explanation alone may introduce additional errors or subtly propagate biases, especially when the model is asked to characterize its own limitations.

\subsection{Path Forward: Building System-Level Interpretability Infrastructure}

The transition from component-level to system-level interpretability requires coordinated action across three dimensions. First, \textbf{technical infrastructure development} must prioritize open-source toolkits that integrate system-level analysis capabilities directly into deployment platforms. Rather than requiring organizations to build custom interpretability systems, default templates should provide system health dashboards, temporal causal analysis, and concurrent execution monitoring as standard features. These tools must support iterative refinement, allowing domain experts to customize system-level explanations for their specific safety requirements without requiring deep interpretability expertise. Synthetic benchmarks with engineered failure modes should be developed to enable controlled validation of system-level techniques before real-world deployment.

Second, \textbf{methodological standardization} must shift evaluation criteria from technical fidelity to operational utility. Success should be measured not by how accurately explanations represent model internals, but by whether they enable stakeholders to predict, prevent, and diagnose failures more effectively than previous approaches. Evaluation frameworks should involve domain experts performing realistic safety oversight tasks using system-level explanations, with metrics capturing whether interventions based on these explanations actually improve system safety outcomes.

Third, \textbf{regulatory evolution} must develop agentic-specific transparency standards that explicitly address temporal causality, multi-component coordination, and emergent behavior. Since system-level properties cannot be satisfied through component-level auditing alone, new frameworks should incentivize holistic safety assurance, demonstrating that integrated systems operate safely across diverse conditions, rather than checkbox compliance showing each component individually passes interpretability tests. Regulatory guidance should specify requirements for temporal accountability (explaining decision sequences, not just individual choices), coordination transparency (revealing multi-agent interaction patterns), and real-time safety monitoring (predictive rather than post-hoc).

\section{Conclusion}
\label{Sec:Conclusion}
This paper has examined the suitability and limitations of existing interpretability methods when applied to agentic systems highlighting key gaps in their ability to capture dynamic, context-dependent, and temporally extended decision processes. Our analysis demonstrates that component-level interpretability, while essential for understanding individual modules, cannot illuminate emergent failures in agentic systems where behavior arises from complex multi-component interactions across time. The research agenda ahead requires developing entirely new frameworks for system-level causal understanding, methods that trace causality across multi-step executions, track error propagation through temporal sequences, translate between heterogeneous explanation types, and maintain attribution clarity under concurrent execution. This is not an incremental improvement to existing techniques but a paradigm shift in how we conceptualize interpretability for autonomous AI. Success will be measured by whether practitioners can use these tools to prevent real failures in deployed systems, ultimately enabling the safe deployment of agentic systems in critical domains where both autonomy and accountability are non-negotiable.

\clearpage
\bibliography{references}

@article{liu2025agentpatterncatalogue,
  title        = {Agent Design Pattern Catalogue: A Collection of Architectural Patterns for Foundation Model Based Agents},
  author       = {Liu, Yue and Lo, Sin Kit and Lu, Qinghua and Zhu, Liming and Zhao, Dehai and Xu, Xiwei and Harrer, Stefan and Whittle, Jon},
  journal      = {Journal of Systems and Software},
  volume       = {220},
  pages        = {112278},
  year         = {2025},
  doi          = {10.1016/j.jss.2024.112278},
  url          = {https://www.sciencedirect.com/science/article/pii/S0164121224003224}
}

@article{sapkota2025ai,
  title        = {{AI} Agents vs.\ Agentic {AI}: A Conceptual Taxonomy, Applications and Challenges},
  author       = {Ranjan Sapkota and Konstantinos I. Roumeliotis and Manoj Karkee},
  journal      = {Information Fusion},
  year         = {2025},
  publisher    = {Elsevier},
  doi          = {10.1016/j.inffus.2025.103599},
  url          = {https://www.sciencedirect.com/science/article/pii/S1566253525006712?via%3Dihub}
}

@article{wang2024survey,
  title        = {A survey on large language model based autonomous agents},
  author       = {Lei Wang and Chen Ma and Xueyang Feng and Zeyu Zhang and Hao Yang and Jingsen Zhang and Zhiyuan Chen and Jiakai Tang and Xu Chen and Yankai Lin and others},
  journal      = {Frontiers of Computer Science},
  volume       = {18},
  number       = {6},
  pages        = {186345},
  year         = {2024},
  publisher    = {Springer},
  doi          = {10.1007/s11704-024-186345-0}
}

@article{futureinternet,
  title        = {The Rise of Agentic {AI}: A Review of Definitions, Frameworks, Architectures, Applications, Evaluation Metrics, and Challenges},
  author       = {Ajay Bandi and Bhavani Kongari and Roshini Naguru and Sahitya Pasnoor and Sri Vidya Vilipala},
  journal      = {Future Internet},
  volume       = {17},
  number       = {9},
  year         = {2025},
  articleno    = {404},
  publisher    = {MDPI},
  doi          = {10.3390/fi17090404},
  url          = {https://www.mdpi.com/1999-5903/17/9/404}
}

@article{elgendy2025agentic,
  title        = {Agentic Systems as Catalysts for Innovation in FinTech: Exploring Opportunities, Challenges and a Research Agenda},
  author       = {Ibrahim A. Elgendy and Mohamed Y. I. Helal and Mohammed A. Al-Sharafi and Mousa Ahmed Albashrawi and Mohammad S. Al-Ahmadi and Il Jeon and Yogesh K. Dwivedi},
  journal      = {Information Discovery and Delivery},
  year         = {2025},
  publisher    = {Emerald},
  doi          = {10.1108/IDD-03-2025-0068}
}

@article{hinostroza2025ai,
  title        = {{AI} with Agency: A Vision for Adaptive, Efficient, and Ethical Healthcare},
  author       = {Vasco Gerardo Hinostroza Fuentes and Hezerul Abdul Karim and Myles Joshua Toledo Tan and Nouar AlDahoul},
  journal      = {Frontiers in Digital Health},
  volume       = {7},
  pages        = {1600216},
  year         = {2025},
  publisher    = {Frontiers Media SA},
  doi          = {10.3389/fdgth.2025.1600216}
}

@article{karunanayake2025next,
  title        = {Next-generation Agentic {AI} for Transforming Healthcare},
  author       = {Nalan Karunanayake},
  journal      = {Informatics and Health},
  volume       = {2},
  number       = {2},
  pages        = {73--83},
  year         = {2025},
  publisher    = {Elsevier},
  doi          = {10.1016/j.infoh.2025.03.001}
}

@article{wei2022chain,
  title        = {Chain-of-Thought Prompting Elicits Reasoning in Large Language Models},
  author       = {Jason Wei and Xuezhi Wang and Dale Schuurmans and Maarten Bosma and Fei Xia and Ed H. Chi and Quoc V. Le and Denny Zhou and others},
  journal      = {Advances in Neural Information Processing Systems},
  volume       = {35},
  pages        = {24824--24837},
  year         = {2022},
  publisher    = {NeurIPS},
  doi          = {10.5555/3600270.3602070}
}

@article{glanois2024survey,
  title        = {A Survey on Interpretable Reinforcement Learning},
  author       = {Claire Glanois and Paul Weng and Matthieu Zimmer and Dong Li and Tianpei Yang and Jianye Hao and Wulong Liu},
  journal      = {Machine Learning},
  volume       = {113},
  number       = {8},
  pages        = {5847--5890},
  year         = {2024},
  publisher    = {Springer},
  doi          = {10.1007/s10994-024-03641-z}
}

@article{murugesan2025rise,
  title        = {The Rise of Agentic {AI}: Implications, Concerns, and the Path Forward},
  author       = {San Murugesan},
  journal      = {IEEE Intelligent Systems},
  volume       = {40},
  number       = {2},
  pages        = {8--14},
  year         = {2025},
  publisher    = {IEEE},
  doi          = {10.1109/MIS.2025.00003}
}

@misc{chatgpt,
  author       = {OpenAI},
  title        = {ChatGPT},
  year         = {2025},
  url          = {https://www.openai.com/chat},
  urldate      = {2025-10-07}
}

@online{Claude,
  author       = {Anthropic},
  title        = {Claude},
  year         = {2025},
  url          = {https://claude.ai},
  urldate      = {2025-10-07}
}

@online{claudecode,
  author       = {Anthropic},
  title        = {Claude Code},
  year         = {2025},
  url          = {https://www.claude.com/product/claude-code},
  urldate      = {2025-10-07}
}

@online{copilot,
  author       = {Microsoft},
  title        = {GitHub Copilot},
  year         = {2025},
  url          = {https://github.com/features/copilot},
  urldate      = {2025-10-07}
}

@misc{hu2023chatgpt,
  author       = {Krystal Hu},
  title        = {Chat{GPT} sets record for fastest-growing user base -- analyst note},
  year         = {2023},
  organization = {Reuters},
  url          = {https://www.reuters.com/technology/chatgpt-sets-record-fastest-growing-user-base-analyst-note-2023-02-01/}
}

@article{kim2025because,
  title={Because we have LLMs, we Can and Should Pursue Agentic Interpretability},
  author={Kim, Been and Hewitt, John and Nanda, Neel and Fiedel, Noah and Tafjord, Oyvind},
  journal={arXiv preprint arXiv:2506.12152},
  year={2025}
}

@article{agrawal2025redefining,
  title        = {Redefining {CX} with Agentic {AI}: Minerva CQ Case Study},
  author       = {Garima Agrawal and Riccardo De Maria and Kiran Davuluri and Daniele Spera and Charlie Read and Cosimo Spera and Jack Garrett and Don Miller},
  journal      = {arXiv preprint arXiv:2509.12589},
  year         = {2025},
  doi          = {10.48550/arXiv.2509.12589},
  url          = {https://arxiv.org/abs/2509.12589}
}

@article{okpala2025modeling,
  title={Agentic AI systems applied to tasks in financial services: modeling and model risk management crews},
  author={Okpala, Izunna and Golgoon, Ashkan and Kannan, Arjun Ravi},
  journal={arXiv preprint arXiv:2502.05439},
  year={2025}
}

@article{derouiche2025agentic,
  title={Agentic ai frameworks: Architectures, protocols, and design challenges},
  author={Derouiche, Hana and Brahmi, Zaki and Mazeni, Haithem},
  journal={arXiv preprint arXiv:2508.10146},
  year={2025}
}

@article{causalsufficiency,
  title        = {Causal Sufficiency and Necessity Improves Chain-of-Thought Reasoning},
  author       = {Xiangning Yu and Zhuohan Wang and Linyi Yang and Haoxuan Li and Anjie Liu and Xiao Xue and Jun Wang and Mengyue Yang},
  journal      = {arXiv preprint arXiv:2506.09853},
  year         = {2025},
  doi          = {10.48550/arXiv.2506.09853},
  url          = {https://arxiv.org/abs/2506.09853}
}

@article{manuvinakurike2025thoughts,
  title        = {Thoughts without Thinking: Reconsidering the Explanatory Value of Chain-of-Thought Reasoning in {LLMs} through Agentic Pipelines},
  author       = {Ramesh Manuvinakurike and Emanuel Moss and Elizabeth Anne Watkins and Saurav Sahay and Giuseppe Raffa and Lama Nachman},
  journal      = {arXiv preprint arXiv:2505.00875},
  year         = {2025},
  doi          = {10.48550/arXiv.2505.00875},
  url          = {https://arxiv.org/abs/2505.00875}
}

@article{chen2025reasoning,
  title        = {Reasoning Models Don't Always Say What They Think},
  author       = {Yanda Chen and Joe Benton and Ansh Radhakrishnan and Jonathan Uesato and Carson Denison and John Schulman and Arushi Somani and Peter Hase and Misha Wagner and Fabien Roger and others},
  journal      = {arXiv preprint arXiv:2505.05410},
  year         = {2025},
  url          = {https://arxiv.org/abs/2505.05410}
}

@article{hassija2024interpreting,
  title        = {Interpreting Black-Box Models: A Review on Explainable Artificial Intelligence},
  author       = {Vikas Hassija and Vinay Chamola and Atmesh Mahapatra and Abhinandan Singal and Divyansh Goel and Kaizhu Huang and Simone Scardapane and Indro Spinelli and Mufti Mahmud and Amir Hussain},
  journal      = {Cognitive Computation},
  volume       = {16},
  number       = {1},
  pages        = {45--74},
  year         = {2024},
  publisher    = {Springer},
  doi          = {10.1007/s12559-023-10167-1}
}

@article{linardatos2020explainable,
  title        = {Explainable {AI}: A Review of Machine Learning Interpretability Methods},
  author       = {Pantelis Linardatos and Vasilis Papastefanopoulos and Sotiris Kotsiantis},
  journal      = {Entropy},
  volume       = {23},
  number       = {1},
  pages        = {18},
  year         = {2020},
  publisher    = {MDPI},
  doi          = {10.3390/e23010018}
}

@article{lundberg2017unified,
  title        = {A Unified Approach to Interpreting Model Predictions},
  author       = {Scott M. Lundberg and Su-In Lee},
  journal      = {Advances in Neural Information Processing Systems},
  volume       = {30},
  pages        = {4765--4774},
  year         = {2017},
  publisher    = {NeurIPS},
  url          = {https://dl.acm.org/doi/10.5555/3295222.3295230}
}

@inproceedings{ribeiro2016should,
  title        = {{``Why Should I Trust You?''} Explaining the Predictions of Any Classifier},
  author       = {Marco Tulio Ribeiro and Sameer Singh and Carlos Guestrin},
  booktitle    = {Proceedings of the 22nd ACM SIGKDD International Conference on Knowledge Discovery and Data Mining},
  pages        = {1135--1144},
  year         = {2016},
  publisher    = {ACM},
  doi          = {10.1145/2939672.2939778}
}

@article{itti2002model,
  title        = {A Model of Saliency-Based Visual Attention for Rapid Scene Analysis},
  author       = {Laurent Itti and Christof Koch and Ernst Niebur},
  journal      = {IEEE Transactions on Pattern Analysis and Machine Intelligence},
  volume       = {20},
  number       = {11},
  pages        = {1254--1259},
  year         = {2002},
  publisher    = {IEEE},
  doi          = {10.1109/34.971829}
}

@inproceedings{selvaraju2017grad,
  title        = {{Grad-CAM}: Visual Explanations from Deep Networks via Gradient-Based Localization},
  author       = {Ramprasaath R. Selvaraju and Michael Cogswell and Abhishek Das and Ramakrishna Vedantam and Devi Parikh and Dhruv Batra},
  booktitle    = {Proceedings of the IEEE International Conference on Computer Vision (ICCV)},
  pages        = {618--626},
  year         = {2017},
  publisher    = {IEEE},
  doi          = {10.1109/ICCV.2017.632}
}

@article{gao2023interpretability,
  title        = {Interpretability of Machine Learning: Recent Advances and Future Prospects},
  author       = {Lei Gao and Ling Guan},
  journal      = {IEEE MultiMedia},
  volume       = {30},
  number       = {4},
  pages        = {105--118},
  year         = {2023},
  publisher    = {IEEE},
  doi          = {10.1109/MMUL.2023.3295405}
}

@article{madsen2401self,
  title={Self-consistency improves chain of thought reasoning in language models},
  author={Wang, Xuezhi and Wei, Jason and Schuurmans, Dale and Le, Quoc and Chi, Ed and Narang, Sharan and Chowdhery, Aakanksha and Zhou, Denny},
  journal={arXiv preprint arXiv:2203.11171},
  year={2022}
}

@article{huang2025survey,
  title        = {A Survey on Hallucination in Large Language Models: Principles, Taxonomy, Challenges, and Open Questions},
  author       = {Lei Huang and Weijiang Yu and Weitao Ma and Weihong Zhong and Zhangyin Feng and Haotian Wang and Qianglong Chen and Weihua Peng and Xiaocheng Feng and Bing Qin and others},
  journal      = {ACM Transactions on Information Systems},
  volume       = {43},
  number       = {2},
  pages        = {1--55},
  year         = {2025},
  publisher    = {ACM},
  doi          = {10.1145/3703155},
}

@article{han2023ignorance,
  title        = {Is Ignorance Bliss? The Role of Post Hoc Explanation Faithfulness and Alignment in Model Trust in Laypeople and Domain Experts},
  author       = {Tessa Han and Yasha Ektefaie and Maha Farhat and Marinka Zitnik and Himabindu Lakkaraju},
  journal      = {arXiv preprint arXiv:2312.05690},
  year         = {2023},
  url          = {https://arxiv.org/abs/2312.05690}
}

@inproceedings{kim2018interpretability,
  title        = {Interpretability Beyond Feature Attribution: Quantitative Testing with Concept Activation Vectors {(TCAV)}},
  author       = {Been Kim and Martin Wattenberg and Justin Gilmer and Carrie Cai and James Wexler and Fernanda Vi{\'e}gas and Rory Sayres},
  booktitle    = {Proceedings of the 35th International Conference on Machine Learning (ICML)},
  pages        = {2668--2677},
  year         = {2018},
  publisher    = {PMLR}
}

@article{lipton2018mythos,
  title        = {The Mythos of Model Interpretability: In Machine Learning, the Concept of Interpretability is Both Important and Slippery},
  author       = {Zachary C. Lipton},
  journal      = {Queue},
  volume       = {16},
  number       = {3},
  pages        = {31--57},
  year         = {2018},
  publisher    = {ACM}
}

@article{guidotti2018survey,
  title        = {A Survey of Methods for Explaining Black Box Models},
  author       = {Riccardo Guidotti and Anna Monreale and Salvatore Ruggieri and Franco Turini and Fosca Giannotti and Dino Pedreschi},
  journal      = {ACM Computing Surveys},
  volume       = {51},
  number       = {5},
  pages        = {1--42},
  year         = {2018},
  publisher    = {ACM}
}
\bibliographystyle{plainnat}


\newpage

\appendix

\section*{Appendix}

\section{Additional Information on AI Agents vs Agentic Systems and their Fundamental Differences} 
\begin{table}[h!]
\centering
\caption{Core Identity, Functionality, Capabilities, and Components}
\label{tab:core_identity}
\renewcommand{\arraystretch}{1.4} 
\small
\begin{tabularx}{\textwidth}{p{0.18\textwidth} X X X}
\toprule
\textbf{Aspect} & \textbf{Large Language Models (LLMs)} & \textbf{AI Agent} & \textbf{Agentic Systems} \\
\midrule
\textbf{Definition} & 
A sophisticated neural network model (e.g., transformer) trained on vast datasets. & 
An overarching software architecture or system built around an LLM. & 
A suite of multiple agents collaborating to solve one higher common complex goal. \\

\textbf{Core Logic} & 
Primarily language generation, comprehension, and pattern recognition. & 
Uses LLM for reasoning, but integrates additional modules for perception, management, and tool use. & 
Uses LLM and core agent modules, as well as more advanced modules in planning, memory, orchestration, and coordination. \\

\textbf{Autonomy} & 
Reactive; responds directly to given prompts, producing output based on internal knowledge. & 
Proactive and autonomous; can initiate actions, iterate through steps, self-correct, and manage multi-step plans. & 
Fully autonomous, capable of solving complex problems collaboratively and influencing other agents dynamically. \\

\textbf{Interaction} & 
Primarily through text input and output; does not inherently interact with external systems beyond generating text. & 
Interacts with the external world (digital or physical) using a suite of tools and APIs, receiving feedback from its actions. & 
Interacts, orchestrates, and plans with multiple sub-agents sequentially or concurrently from input to output generation. \\

\textbf{Memory} & 
Limited by its context window; does not maintain persistent memory beyond current interaction unless externally managed. & 
Incorporates limited short-term (contextual) and long-term (persistent knowledge base, vector databases) memory components. & 
Has persistent, multi-agent memory, including short-term (working memory) and long-term (episodic and semantic memory). \\

\textbf{Learning} & 
Learning occurs during pre-training; relies on fine-tuning for adaptation. & 
Can exhibit continuous learning through feedback loops, refining strategies and knowledge based on outcomes. & 
Continuously learns and adapts to new situations by adjusting goals and coordination strategies dynamically. \\

\textbf{Example Task} & 
Answering “What is the capital of France?”, summarizing an article, writing a poem. & 
“Book a flight from London to New York for next Tuesday, then find a highly rated restaurant near the airport.” & 
A customer support system that uses multiple agents (e.g., for communication, research) to detect anomalies, initiate actions, and resolve issues end-to-end without human intervention. \\
\bottomrule
\end{tabularx}
\end{table}

\section{Additional Notes on Limitations of Explaining Agentic Systems} 

A number of popular post-hoc interpretability methods have been developed to probe black-box models’ decisions, but each has significant limitations when applied to agentic systems. Post-hoc methods generally provide after-the-fact explanations (e.g. feature importance or visualizations) for a model’s output without requiring the model to be intrinsically transparent. While these can be useful for single-model settings, their assumptions break down in the context of complex multi-agent reasoning:

  \paragraph{LIME (Local Interpretable Model-Agnostic Explanations):} LIME explains an individual prediction by training a simple surrogate (like a linear model) around that instance. This works for isolated decisions, but in an agentic system an “instance” may be an entire sequence of interactions rather than a fixed feature vector. Moreover, LIME’s local linear approximation means it fails to capture non-linear feature interactions in the true model’s behavior (paper). LIME is also limited to local explanations; it cannot provide a global view of an agent’s strategy over many states or how multiple agents’ behaviors combine. These limitations make LIME’s explanations fragmentary and potentially unreliable for understanding an agentic system’s behavior.

  \paragraph{Attention/Saliency Maps:} For models like transformers, one might attempt to use attention weights or gradient-based saliency as explanations (e.g. highlighting which words or state elements an agent “focused” on). This, too, has limited utility in agentic systems. In a multi-agent LLM system, an agent’s policy might not even expose attention weights to the end-user, and even if it does, making sense of dozens of attention heads over a long dialogue between agents is intractable for a human. Similarly, gradient or perturbation saliency methods (which highlight parts of the input that most affect the output) struggle when the “input” is an entire history of an agent’s perceptions or messages. These methods produce heatmaps or importance scores that are hard to translate into a high-level explanation. In practice, saliency maps tend to be too low-level and instance-specific to explain an agent’s behavior to managers – they might show which token changed an output, but not why the agent chose a particular strategy in an interactive context.

  \paragraph{Chain of Thought Reasoning: }Research reconsidering the explanatory value of CoT through agentic pipelines reveals fundamental limitations in using CoT verbalization as a reliable interpretability mechanism. \cite{manuvinakurike2025thoughts} demonstrate that in agentic pipeline contexts, CoT reasoning does not lead to better outputs nor does it offer genuine explainability, instead producing seemingly plausible but often erroneous, contradictory, or irrelevant explanations that fail to improve end users' ability to understand systems or achieve their goals. These concerns are substantiated by empirical evidence showing that reasoning models' CoTs often lack faithfulness to their actual decision-making processes. \cite{chen2025reasoning} found that state-of-the-art reasoning models verbalize the reasoning factors they actually use in only 25-39\% of cases, with faithfulness dropping to 20-29\% for potentially misaligned behaviors. Particularly concerning is evidence that models can generate elaborate justifications while concealing problematic reasoning, with some reward hacking behaviors verbalized in fewer than 2\% of cases despite being exploited consistently. These findings suggest that CoT verbalizations may not faithfully represent the computational processes underlying model decisions, limiting their utility for understanding system-level behavior in agentic systems where decisions propagate across multiple steps and components.



\end{document}